\documentclass{article}

\usepackage{microtype}
\usepackage{graphicx}
\usepackage{subfigure}
\usepackage{booktabs} 

\usepackage{hyperref}

\usepackage{iclr2026/iclr2026_conference, times}
\iclrfinalcopy

\usepackage{amsmath}
\usepackage{amssymb}
\usepackage{mathtools}
\usepackage{amsthm}

\usepackage[capitalize,noabbrev]{cleveref}

\theoremstyle{plain}
\newtheorem{theorem}{Theorem}[section]
\newtheorem{proposition}[theorem]{Proposition}

\theoremstyle{definition}
\newtheorem{definition}[theorem]{Definition}

\theoremstyle{remark}

\usepackage{mathrsfs}
\usepackage{bm}

\usepackage[textsize=tiny]{todonotes}

\title{From Neural Networks to Logical Theories: The Correspondence between Fibring Modal Logics and Fibring Neural Networks}

\author{
  Ouns El Harzli\textsuperscript{1,2, 3}\thanks{Correspondence to: \texttt{ouns.elharzli@new.ox.ac.uk}} \quad
  Bernardo Cuenca Grau \textsuperscript{1} \quad
  Artur D'Avila Garcez\textsuperscript{4} \AND  \begin{tabular}{r}
    Ian Horrocks\textsuperscript{1} \quad
    Tarek R. Besold\textsuperscript{2}
  \end{tabular} \AND
  \quad \\
  \begin{tabular}{l}
  \textsuperscript{1}Department of Computer Science, University of Oxford, UK\\
  \textsuperscript{2}Sony AI Barcelona, Spain\\
  \textsuperscript{3}Sony AI Tokyo, Japan\\
  \textsuperscript{4}City St George's, University of London, UK
  \end{tabular}
}

\begin{document}

\maketitle

\begin{abstract}
    Fibring of modal logics is a well-established formalism for combining countable families of modal logics into a single fibred language with common semantics, characterized by fibred models. Inspired by this formalism, fibring of neural networks was introduced as a neurosymbolic framework for combining learning and reasoning in neural networks. Fibring of neural networks uses the (pre-)activations of a trained network to evaluate a fibring function computing the weights of another network whose outputs are injected back into the original network. However, the exact correspondence between fibring of neural networks and fibring of modal logics was never formally established. In this paper, we close this gap by formalizing the idea of fibred models \emph{compatible} with fibred neural networks. Using this correspondence, we then derive non-uniform logical expressiveness results for Graph Neural Networks (GNNs), Graph Attention Networks (GATs) and Transformer encoders. Longer-term, the goal of this paper is to open the way for the use of fibring as a formalism for interpreting the logical theories learnt by neural networks with the tools of computational logic. 
\end{abstract}

\section{Introduction}

The advent of large language models has created unprecedented interest in the task of reasoning in neural networks. Logical reasoning is arguably the best perspective to study and develop this capability, offering precise definitions, validity conditions and a formalism that is amenable to formal verification. As a result, there has been a surge of interest in the field of neurosymbolic AI that studies the integration of neural networks with logical reasoning \citep{Besold2022,Lamb2020,GarcezLamb2023}.

Fibring of Neural Networks \citep{10.5555/1597148.1597205} is a theoretical concept from the neurosymbolic AI literature introduced as a way of combining neural network architectures. The idea is to enforce that the parameters of a network are a function of another network, and that the resulting output is injected back into the original network when computing the output of the combined (fibred) neural network. This theoretical framework was initially inspired by the concept of fibring logics \citep{Gabbay1999-GABFL}, in particular combining modal logics \citep{chellas1980modal}, which play an important role in systems verification since temporal logic, a special case of modal logic, is extensively used in verification. Fibred neural networks were shown in \citep{10.5555/1597148.1597205} to be strictly more expressive than the usual composition of neural networks. Fibred networks intended to offer a framework for the study of the combination of learning and reasoning in neural networks, whereby one network's learning influences another network's inference. 

In a fibred modal language, the Kripke model and the world in which a formula is evaluated are a function of a possible world in another Kripke model.\footnote{The word \textit{model} here refers to assignments of truth-values, differently from the use of the word in neural networks.} However, the precise correspondence between fibring of neural networks and fibring of modal logics was never formally established.

An important research area in neurosymbolic AI aims to establish connections between logic and modern neural architectures, motivated by the prospect of rendering the latter more interpretable and verifiable. In particular, Graph Neural Networks (GNNs) and Transformers have become essential components in contemporary machine learning, each tackling specific but sometimes intersecting challenges across a wide range of applications. GNNs excel at processing structured graph data, finding extensive use in fields such as social network analysis, drug discovery, and knowledge graphs \citep{SALAMAT2021106817,xiong2021graph,ye2022comprehensive}. Transformers have revolutionized natural language processing by modeling intricate contextual relationships in sequential data, providing unprecedented capabilities for tasks like language understanding and generation \citep{NIPS2017_3f5ee243,devlin2019bert,brown2020language,openai2023gpt4}. The capabilities of these architectures have sparked a significant research effort to rigorously analyze their \emph{logical expressiveness}--i.e., the classes of formal languages they can compute--and their formal verification. An interesting connection exists between GNNs and Transformer encoders \citep{DBLP:journals/corr/abs-2104-13478}: Transformer encoders can be viewed as GNNs applied to complete graphs with added positional encoding and an attention mechanism. This connection, however, has not been exploited in the literature on logical expressivity and formal verification. This gap leaves potential for unifying logical characterizations of both architectures.

Existing logical expressiveness results for GNNs and Transformers can be categorized into: \emph{discriminative power} (e.g.\ \textit{is there a logical classifier that can distinguish the same pairs of nodes as a GNN?}); \emph{uniform expressiveness} (e.g.\, \textit{is there a logical formula whose truth values coincide with the output of a GNN given \emph{any} input graph and node?}); and \emph{non-uniform expressiveness} 
(e.g.\ \textit{is there a logical formula whose truth values coincide with the output of a GNN depending on the input?}). In this categorization, uniform expressiveness stand out as the most powerful kind, providing complete logical characterizations for neural architectures (i.e.\ providing a single logical formula for each instance of a neural architecture independently of the input). Initial results established that the discriminative power of standard GNNs is upper-bounded by the Weisfeiler-Leman (WL) test \citep{barcelo2020logical}. Subsequently, \cite{Grohe2023} established non-uniform expressiveness results for GNNs in terms of Boolean circuits and descriptive complexity. More recently, uniform expressiveness results of broad classes of GNNs have been derived \citep{ijcai2024p391,benedikt_et_al:LIPIcs.ICALP.2024.127}, using known logics called Presburger logics, which involve counting modalities and linear inequalities. Using existing results about the satisfiability of such logics, results about the computational complexity of formal verification problems for networks follow as corollaries (e.g.\ answering the question: \textit{given an output and a GNN, is there an input to the GNN that yields that output?}). In particular, \cite{benedikt_et_al:LIPIcs.ICALP.2024.127} provides a taxonomy for the decidability of these verification problems depending on the types of aggregations and activation functions. Then, \cite{grau2025correspondenceboundedgraphneural} established uniform expressiveness results for bounded GNNs in terms of fragments of first-order logic, showcasing how restrictive yet practical assumptions on the class of GNNs may simplify significantly their logical characterization. Transformer encoders have also been studied through the lens of circuit complexity theory. Unique Hard Attention Transformers (UHATs) have been mapped to fragments of the complexity class of AC$^0$ languages with extensions based on a restricted form of first-order logic \citep{hao2022formal}. Average Hard Attention Transformers (AHATs) have been shown capable of capturing more complex languages, including those outside AC$^0$, and most recently, a (uniform) lower bound has been established in terms of linear temporal logic: a language called LTL (C, +), which also involves counting modalities \citep{barcelo2024logical}.

In this paper, we propose fibring of modal logics as a new formalism to study the logical expressiveness of neural architectures, including GNNs and Transformers. We start by redefining fibring of neural networks \citep{10.5555/1597148.1597205} to make its use easier in the context of the above literature. We then formally establish an exact correspondence between fibring neural networks and fibring modal logics. That is, (i) we define a fibred language based on a fibring architecture; (ii) we introduce the notion of \emph{compatible} fibred models (with fibred neural networks and their inputs) on which to interpret the formulas; (iii) we prove that our proposed fibred logic is a valid fibred logic by demonstrating that the class of compatible fibred models is non-empty and closed under Kripke-model isomorphism; and (iv) we construct the formulas from our fibred logic whose truth values coincide with the outputs of fibred neural networks. Subsequently, we prove that fibred neural networks can be used to non-uniformly describe large classes of GNNs, GATs and Transformer encoder architectures. It follows that the corresponding fibred languages provide non-uniform expressiveness results for these network architectures. Our hope is that fibring can become a unifying formalism for the study of GNNs and Transformers in neurosymbolic AI.

As future directions of research, we speculate that fibring as a formalism might hold the key to deriving and unifying uniform expressiveness results for GNNs and Transformer encoders, and we present our main arguments for this possible unification in \Cref{sec:discussion}. We argue that the proposed fibring framework provides a new intuitive way of thinking about GNNs and Transformer architectures as successive combinations of underlying logics, and we discuss how this new perspective could be used for interpretability and verification.

The paper is organized as follows. \Cref{sec:prelim} introduces the notation used throughout the paper. \Cref{fnn} provides the new definition of fibring. \Cref{sec:equivalence} proves the correspondence between fibring networks and modal logics, and \Cref{sec:expressivity} applies this result to derive expressiveness results for GNNs, Graph Attention Networks (GATs) and Transformer Encoders. \Cref{sec:discussion} concludes the paper and discusses directions for future work. 






\section{Preliminaries}\label{sec:prelim}

We work with vectors and matrices of rational numbers.
The $i$th entry of a vector $\mathbf{v}$ is written $v_i$,
and the entry in row $i$, column $j$ of a matrix $\mathbf{A}$
is written $A_{i,j}$.
If two matrices have the same number of columns,
we can place them side by side to form a larger matrix,
called their concatenation.

A \emph{neural architecture} $\mathbb{A}$ is specified by the number of its layers $L$,
the dimension of each layer $d_\ell$, and an activation map $\sigma^\ell: \mathbb{Q}^{d_\ell} \mapsto \mathbb{Q}^{d_\ell}$ for each hidden layer. With $L$ layers, $d_0$ and $d_L$
are referred to as the input and output dimensions, respectively,
and for each hidden layer $\ell$, the activation map
$\sigma^\ell$ is computable in polynomial time.
We may also talk about a portion of the architecture
ranging from layer $p$ to layer $q$,
and call this the \emph{sub-architecture} $\mathbb{A}^{p:q}$.
If $q=L$, we write $\mathbb{A}^{p\uparrow}$ for simplicity.

A \emph{network instance} of an architecture assigns weights and biases, known as the set of parameters, to each layer.
The weights of layer $\ell$ form a matrix $\mathbf{W}^{\ell} \in \mathbb{Q}^{d_{\ell} \times d_{\ell-1}}$,
and the biases form a vector $\mathbf{b}^{\ell} \in \mathbb{Q}^{d_{\ell}}$.
To apply a network instance $\mathcal{N}$ to an input vector $x$,
we compute the weighted sum of the input weighted by the parameters followed by the application of the activation function in the usual way proceeding from the input layer to the output.
The final output of the network is a vector of the output's dimension.
Formally, the application of $\mathcal{N}$ to $\mathbf{x} \in \mathbb{Q}^{d_0}$ generates a sequence
$\mathbf{h}^{1}, \ldots, \mathbf{h}^{L}$ of vectors defined as $\mathbf{h}^{\ell}  =   \mathbf{W}^{\ell} \cdot \mathbf{x}^{\ell-1} + \mathbf{b}^{\ell}$, where $\mathbf{x}^{0} = \mathbf{x}$,
$\mathbf{x}^{\ell} = \sigma^{\ell}(\mathbf{h}^{\ell})$.
The result $\mathcal{N}(\mathbf{x})$ of applying $\mathcal{N}$ to  $\mathbf{x}$ is the vector $\mathbf{h}^{L}$. 
We also talk about \emph{sub-networks}.
The sub-network $\mathcal{N}^{p:q}$ consists of the layers
from $p$ through $q$ and parameters.
If $q=L$, we write $\mathcal{N}^{p\uparrow}$.

\section{Fibring Neural Networks}\label{fnn}

In this section, we introduce a definition of fibred neural networks which generalizes the original definition of \citet{10.5555/1597148.1597205} to any number and possible combinations of neural networks.

A \emph{fibring architecture} is a directed tree $\mathcal{F}$ whose nodes $v$
are labeled with neural architectures $\mathbb{A}_v$.
Each edge $(v,v')$  is labeled with: (i) a layer number $\ell$ in the parent architecture; and (ii) a set of positions $S$ in that layer denoting the fibred neurons. We impose the additional requirement that for any two edges $(v,w)$ and $(v,u)$ sharing the parent node and labeled with the same layer number, the corresponding sets of positions must be disjoint. Finally, we denote by $\mathscr{F}$ the class of fibring architectures which verify the property that: (i) the architecture at the root has two linear layers with input dimension $n$ and output dimension $1$, and (ii) every edge leaving the root node is labeled with the same layer number $\ell$.

A \emph{fibred network} $\widetilde{\mathcal{N}}$ is a tuple $\langle \mathcal{N}, \mathcal{F}, \Tilde{f}\rangle$ where $\mathcal{N}$ is a network instance, $\mathcal{F}$ is a fibring architecture, and $\Tilde{f}$ is a finite collection of neural \emph{fibring functions} matching $\mathcal{F}$, i.e.\ one function $\Tilde{f}_{(v, v')}$ for each edge $(v,v')$ in $\mathcal{F}$, which specifies how to build
an instance of the child architecture and what input to give it; specifically, for an edge $(v,v')$ labeled with layer number $\ell$, the fibring function $\Tilde{f}_{(v, v')}$ maps vectors of
dimension $\leq d_{\ell}$ to a network instance of architecture $\mathbb{A}_{v'}$ and a valid input vector for that architecture.

To apply a fibred neural network $\widetilde{\mathcal{N}}$ to an input $\mathbf{x}$, we start at the root.
Whenever we reach a layer that has edges leading to children,
we pause, call the corresponding fibring function, and pass part of the current vector
into the child network.
The child network produces a new vector,
which is spliced back into the parent’s computation
at the specified positions.
The overall computation proceeds recursively until the output layer
of the root is reached.

Formally, the computation is defined inductively.
If $\mathcal{F}$ has only the root node $u$,
then $\widetilde{\mathcal{N}}(\mathbf{x}) = \mathcal{N}(\mathbf{x})$.
Otherwise, let $u_1,\dots,u_k$ be the children of $u$,
with edges $(u,u_i)$ labeled $(\ell_i, S_i)$, let $\mathcal{F}_i$ be the subtree rooted at $u_i$, and let $\Tilde{f}_i$ be the restriction of $\Tilde{f}$ to edges in $\mathcal{F}_i$.
Assume the children are ordered so that
$\ell_1 \leq \ell_2 \leq \cdots \leq \ell_k$.
For each stage $1 \leq i \leq k$, define a tuple
$(\mathbf{x}_i, \mathcal{N}_i, \mathbf{y}_i, \mathbf{h}_i)$ as follows:
\begin{align*}
\mathbf{x}_i &=
\begin{cases}
\mathcal{N}^{1:\ell_1}(\mathbf{x}), & i=1,\\[2pt]
\mathcal{N}^{\ell_{i-1}:\ell_i}(\mathbf{y}_{i-1}), & i>1,
\end{cases}\\
(\mathcal{N}_i,\mathbf{y}_i) &= \Tilde{f}_{(u,u_i)} (\mathbf{x}_i),\\
\mathbf{h}_i &= \text{$\mathbf{x}_i$ with entries in $S_i$
replaced by } \langle \mathcal{N}_i,\mathcal{F}_i, \Tilde{f}_i\rangle(\mathbf{y}_i), \text{the application of the fibred network to } \mathbf{y}_i.
\end{align*}

The final output is:
\[
\widetilde{\mathcal{N}}(\mathbf{x}) \;=\; \mathcal{N}^{\ell_k\uparrow}(\mathbf{h}_k).
\]

If the output of the root network is a scalar, we can interpret the fibred network as a classifier in the usual way:
on input $\mathbf{x}$, it outputs \emph{true} if the final value is strictly greater than $0$,
and \emph{false} otherwise.

\section{Exact correspondence between fibring logics and fibring networks}\label{sec:equivalence}

In this section, we introduce a fibred modal language for combining different modal systems
and show it captures exactly the behavior of fibred neural networks.

\paragraph{Modal logics and fibring \citep{Gabbay1999-GABFL}:}\label{modal logics} 
Fix a finite set of propositions $PROP = \{p_1,\dots,p_n\}$ and a countable
collection of modal operators $\Box_i$ (one for each index $i$).
For each $i$, logic $\mathcal{L}_i$ has formulas built from propositions,
$\top$ (true), Boolean connectives, and the modal operators $\Box_i$.
The semantics is the usual Kripke semantics:
each $\mathcal{L}_i$ uses a class of Kripke models,
and $\Box_i \varphi$ holds at a world $w$ in a model when $\varphi$ holds at all accessible worlds $w_i$ from $w$ according to a pre-defined accessibility relation $R$ such that $R(w_i,w)$ holds.

To \emph{fibre} these logics, we allow all the modal operators $\Box_i$
in one combined language.
That is, formulas are built from the grammar:
\begin{align}
\varphi ::= p \mid \top \mid \varphi_1 \wedge \varphi_2 \mid \neg \varphi \mid \Box_i\varphi
\end{align}
where $p \in PROP$ and $i$ ranges over the fibred modal logics $\mathcal{L}_i$.

A \emph{fibred model} chooses,
for each $i$, one Kripke model for $\mathcal{L}_i$,
and also provides a way of \emph{jumping} from worlds of other models
into the model for $i$.
Intuitively, evaluating $\Box_j \varphi$ at a world in model $i$
either uses the native accessibility for the model $i$ if $j=i$,
or first jumps into model $j$ and then evaluates there.\footnote{
We assume that the classes of models are disjoint so that each world’s \textit{home} model is unambiguous and we also assume that all component Kripke models are finite.}
Formally, a fibred model $\mathcal{M}$ consists of one Kripke model $\mathbf{m}_i$
for each logic $\mathcal{L}_i$, together with a family of logical
\emph{fibring functions} $f_i$.
Each $f_i$ maps each world in another model $\mathbf{m}_j$ ($j \neq i$)
to a world in $\mathbf{m}_i$, while $f_i$ tells us how to jump from another component into the $i$-th one.

Satisfaction in a fibred model $\mathcal{M}$ is defined as follows.
If $w$ is a world in $\mathbf{m}_i$, then:

\begin{itemize}
\item $\mathcal{M},w \models p$ iff $w$ is in the valuation of $p$ in $\mathbf{m}_i$.
\item $\mathcal{M},w \models \top$ always.
\item $\mathcal{M},w \models \varphi_1 \wedge \varphi_2$ iff
      $\mathcal{M},w \models \varphi_1$ and $\mathcal{M},w \models \varphi_2$.
\item $\mathcal{M},w \models \neg \varphi$ iff not $\mathcal{M},w \models \varphi$.
\item $\mathcal{M},w \models \Box_j\varphi$ iff \emph{(i)} $j=i$ and for all $w'$ with $(w,w')$ in the accessibility relation of $\mathbf{m}_i$, we have $\mathcal{M},w' \models \varphi$; or \emph{(ii)} $j \neq i$ and $\mathcal{M},f_j(w) \models \Box_j \varphi$.
\end{itemize}

\paragraph{A fibred logic for a fibring architecture:}\label{fragmentlogic}

Let $\mathcal{F}$ be a fibring architecture.
For each node $v$ in $\mathcal{F}$
and each layer number $\ell$ of $\mathbb{A}_v$, associate a distinct modal
operator $\Box_{v,\ell}$, interpreted over the class of all finite Kripke models; let $\mathcal{L}$ be the resulting fibred logic.
Based on $\mathcal{L}$ we will define another fibred logic $\mathcal{L}_{\mathcal{F}}$ using the notion
of \textit{compatible} models, defined next.

\begin{definition}

Let $\mathbf{m}$ be a Kripke model and $\mathcal{N}$ a network instance.
A map $\pi$ that assigns to each world of $\mathbf{m}$ an input
vector for $\mathcal{N}$ is \emph{$(\mathbf{m},\mathcal{N})$–admissible} if it is injective and, for
all worlds $w,w'$,
\[
w \text{ is related to } w' \text{ in } \mathbf{m}
\quad\Longleftrightarrow\quad
\mathcal{N}\big(\pi(w)\big)=\mathcal{N}\big(\pi(w')\big).
\]

\end{definition}

\begin{definition}
Let $\mathcal{M}$ be a fibred model of $\mathcal{L}$
and denote with $\mathbf{m}_{v,\ell}$ the component Kripke model in $\mathcal{M}$
for node $v$ in $\mathcal{F}$ and layer number $\ell$ in $\mathbb{A}_v$.
Let
$\widetilde{\mathcal{N}}=\langle \mathcal{N},\mathcal{F}, \Tilde{f}\rangle$, let $\mathbf{x} \in \{0,1\}^n$, and let $u$ be the root of $\mathcal{F}$.
We say that $\mathcal{M}$ is \emph{compatible} with
$(\widetilde{\mathcal{N}}, \mathbf{x})$ if we can assign:
\begin{itemize}
 \item  to each node $v$ in $\mathcal{F}$, a network instance $\mathcal{N}_v$ of $\mathbb{A}_v$, and
 \item  to each pair $(v, \ell)$ of a node in $\mathcal{F}$ and a layer $\ell$ in $\mathbb{A}_v$, a bijection $\pi_{v, \ell}$ from a finite set of Kripke worlds to a finite set of vectors,
\end{itemize}
 such that the following compatibility conditions hold: 

\begin{itemize}
\item[(C0)] $\mathcal{N}_u = \mathcal{N}$ and  $\pi_{u,1}$ maps each world $w$ in $\mathbf{m}_{u,1}$  to the vector in $\{0,1\}^n$ whose $i$-th bit
is $1$ exactly when proposition $p_i$ is true at $w$.


\item[(C1)] $\pi_{v,\ell}$
is $(\mathbf{m}_{v, \ell}, \mathcal{N}_v^{\ell\uparrow})$-admissible.


\item[(C2)] Assume $v$ has children $v_1,\dots,v_k$, ordered by the layer numbers $\ell_1, ..., \ell_k$ labeling edges $(v, v_i)$, and let $w$
be the world reached via the composition of logical fibring functions from $(\pi_{u,1})^{-1} (\mathbf{x})$ to Kripke model $\mathbf{m}_{v,1}$.
Running $\langle \mathcal{N}_v,\mathcal{F}_v, \Tilde{f}_v\rangle$ on $\pi_{v,1} (w)$
produces tuples $(\mathbf{x}_i,\mathcal{N}_i,\mathbf{y}_i,\mathbf{h}_i)$ for $1 \leq i \leq k$.
Then:
\begin{enumerate}
\item $\mathcal{N}_{v_i}=\mathcal{N}_i$
and $\mathbf{y}_i=\pi_{{v_i}, 1} (f_{v_i, 1}(w))$.
\item $\mathbf{h}_i = \pi_{v, \ell_i} (f_{v, \ell_i}(w))$.
\item The set of propositions that are true at world $\pi_{v,\ell_k}^{-1}(\mathbf{h}_k)$ in model $\mathbf{m}_{v, \ell_k}$
agrees with the set of propositions that are true at $\pi_{{v_i},1}^{-1} (\mathbf{y}_i)$  in model $\mathbf{m}_{{v_i}, 1}$ for $1 \leq i \leq k$.
\end{enumerate}

\end{itemize}
\end{definition}

For a fixed pair $(v,\ell)$ of a node $v$ in $\mathcal{F}$ and a layer number $\ell$ in $\mathbb{A}_v$, let:
\[
\mathsf{Comp}_{\mathcal{F}}(v,\ell)
\ :=\ \left\{\ \mathbf{m}\ \middle|\ 
\begin{array}{l}
\exists\,\widetilde{\mathcal{N}}, \mathbf{x}, \mathcal{M}\text{ a fibred model of } \mathcal{L} \text{ compatible with } (\widetilde{\mathcal{N}}, \mathbf{x})\\
\text{such that the $(v, \ell)$-component of }\mathcal{M}\text{ is }\mathbf{m}
\end{array}\right\}.
\]
In words, $\mathsf{Comp}_{\mathcal{F}}(v,\ell)$ is the \emph{projection} of all
compatible fibred models onto their $(v,\ell)$ component.

\begin{proposition}\label{prop:class-compatible}
The class $\mathsf{Comp}_{\mathcal{F}} (v,\ell)$
is non-empty and closed under Kripke-model isomorphism.
\end{proposition}

\emph{Proof.} \emph{Non-empty.} The class $\mathsf{Comp}_{\mathcal{F}}(v,\ell)$ contains at least the Kripke models of fibred models obtained by explicitly running the computation of fibred networks $\Tilde{\mathcal{N}}$ on vectors $\mathbf{x}$.

Specifically, for each $\Tilde{\mathcal{N}} = \langle \mathcal{N}, \mathcal{F}, \Tilde{f}\rangle$ and $\mathbf{x}$, we can construct a compatible fibred model as follows.

We start from the root model: the domain of the root model contains a distinct element $w_\mathbf{z}$ for each vector $\mathbf{z} \in \{0,1\}^n$, two worlds $w_\mathbf{z}, w_{\mathbf{z}'}$ are related iff $\mathcal{N} (\mathbf{z}) = \mathcal{N} ({\mathbf{z'}})$, and the valuation is given by: $p_j$ holds at world $w_\mathbf{z}$ in the root model iff $z_j = 1$.

We run the fibred neural network $\Tilde{\mathcal{N}}$ on all vectors $\mathbf{z} \in \{0,1\}^n$ (recursively generating sequences of tuples $(\mathbf{x}_i,\mathcal{N}_i,\mathbf{y}_i,\mathbf{h}_i)$) and we collect at each $(v,\ell)$ all vectors attained by the computations, that is, for each $(v, \ell_i)$ we collect vectors $\mathbf{h}_i$ and for each $(v_i, 1)$, we collect vectors $\mathbf{y}_i$. Furthermore, we run the fibred neural network $\Tilde{\mathcal{N}}$ on the specific vector $\mathbf{x}$ and we collect at each $v_i$, the network instance $\mathcal{N}_i$ attained by the computation.

We then define Kripke models as follows: the domain of $\mathbf{m}_{v,\ell}$ contains a distinct element $w_\mathbf{v}$ for each vector $\mathbf{v}$ collected at $(v, \ell)$, two worlds $w_\mathbf{v}, w_{\mathbf{v}'}$ are related iff $\mathcal{N}_v^{\ell \uparrow} (\mathbf{v}) = \mathcal{N}_v^{\ell \uparrow} ({\mathbf{v'}})$ where $\mathcal{N}_v$ is the network instance collected at $v$. 

The valuations are defined recursively starting from the root model: a proposition $p$ holds at world $w_\mathbf{h}$ in model $\mathbf{m}_{v, \ell_i}$ iff there is a world $w_\mathbf{v}$ in $\mathbf{m}_{v, 1}$ such that $\mathbf{h} = \mathbf{h}_i$ is attained by the computation of $\langle \mathcal{N}_v, \mathcal{F}_v, \Tilde{f}_v \rangle (\mathbf{v})$ and $p$ holds at $w_\mathbf{v}$ in model $\mathbf{m}_{v, 1}$; and a proposition $p$ holds at world $w_\mathbf{y}$ in model $\mathbf{m}_{{v_i}, 1}$ iff there is a world $w_\mathbf{v}$ in $\mathbf{m}_{v,1}$ such that $\mathbf{y} = \mathbf{y}_i$ is attained by the computation of $\langle \mathcal{N}_v, \mathcal{F}_v, \Tilde{f}_v \rangle (\mathbf{v})$ and $p$ holds at $w_{\mathbf{h}_k}$ in model $\mathbf{m}_{v, \ell_k}$. 

Finally, we require that the (logical) fibring function from $\mathbf{m}_{v, 1}$ to $\mathbf{m}_{v, \ell_i}$ verifies for each world $w_\mathbf{v}$ in $\mathbf{m}_{v, 1}$, $f_{v, \ell_i} (w_\mathbf{v}) = w_{\mathbf{h}_i}$ where $\mathbf{h}_i$ is the vector attained by the computation of $\langle \mathcal{N}_v, \mathcal{F}_v, \Tilde{f}_v \rangle (\mathbf{v})$; and the (logical) fibring function from $\mathbf{m}_{v, 1}$ to $\mathbf{m}_{{v_i}, 1}$ verifies, for each world $w_\mathbf{v}$ in $\mathbf{m}_{v, 1}$, $f_{{v_i}, 1} (w_\mathbf{v}) = w_{\mathbf{y}_i}$ where $\mathbf{y}_i$ is the vector attained by the computation of $\langle \mathcal{N}_v, \mathcal{F}_v, \Tilde{f}_v \rangle (\mathbf{v})$.

The above proof construction covers all compatibility conditions and ensures that the resulting fibred model is compatible with $(\Tilde{\mathcal{N}}, \mathbf{x})$ by taking the bijections $\pi_{v, \ell}: w_\mathbf{v} \mapsto \mathbf{v}$.

\emph{Closed under Kripke-model isomorphism.} Let $\mathbf{m}\in\mathsf{Comp}_{\mathcal{F}} (v,\ell)$ come from some compatible
$\mathcal{M}$, and let $\pi:\mathbf{m}\cong\mathbf{m}'$ be an isomorphism.
Form a new fibred model $\mathcal{M}'$ by replacing the
$(v, \ell)$-component of $\mathcal{M}$ with $\mathbf{m}'$ and replace the bijection
$\pi_{v,\ell}$ by $\pi_{v,\ell}\circ\pi^{-1}$.
This preserves (C0)–(C2): edges, valuations, and fibring jumps are transported
through $\pi$ without changing any observable behavior. Hence
$\mathcal{M}'$ is still compatible with the same $(\widetilde{\mathcal{N}},\mathbf{x})$,
so $\mathbf{m}'\in\mathsf{Comp}_{\mathcal{F}} (v,\ell)$. \qedsymbol

 This shows that the subclass of compatible models is a valid fibred logic, since
 the projections of compatible models represent a non-empty class of Kripke models closed under isomorphism (hence a reasonable class of Kripke models).
 We refer to this fibred logic as $\mathcal{L}_{\mathcal{F}}$.

\paragraph{Equivalence between fibring neural networks and a fragment of fibred logic:}

\begin{definition}\label{def:syntactic-fragment}
Let $ \mathcal{F}$ be a fibring architecture and let $\varphi$ be a propositional formula.
The formula $\psi(\varphi,\mathcal{F}) \in \mathcal{L}_{\mathcal{F}}$ is recursively defined as follows:
\begin{itemize}
\item If $\mathcal{F}$ has only a single node,
then $\psi(\varphi,\mathcal{F})=\varphi$.
\item Otherwise, let $v_1,\dots,v_k$ be the children of the root,
with corresponding subtrees $\mathcal{F}_1,\dots,\mathcal{F}_k$.
For each $i$, let $l_i$ be the largest layer label on an edge leaving $v_i$ (or $l_i=1$ if none).
Then set
\[
\psi(\varphi,\mathcal{F})
= \Box_{{v_1},l_1}\psi(\varphi,\mathcal{F}_1)
  \;\wedge\; \cdots \wedge\;
 \Box_{{v_k},l_k}\psi(\varphi,\mathcal{F}_k).
\]
\end{itemize}

\end{definition}

\begin{theorem}\label{fnneqlogic}
Let $\mathcal{F} \in \mathscr{F}$ be a fibring architecture rooted at $u$. For every network instance $\mathcal{N}$ of the root architecture $\mathbb{A}_u$, there is a propositional formula $\varphi$
such that, for every input $\mathbf{x}\in\{0,1\}^n$, for every collection of fibring functions $\Tilde{f}$ matching $\mathcal{F}$,
and every fibred model $\mathcal{M}$ compatible with
$\langle \mathcal{N},\mathcal{F}, \Tilde{f}\rangle$ and $\mathbf{x}$,
we have that the following holds:
\[
\mathcal{M},(\pi_{u,1})^{-1}(\mathbf{x}) \models \psi(\varphi,\mathcal{F})
\quad\text{iff}\quad  \langle \mathcal{N},\mathcal{F}, \Tilde{f}\rangle \text{ classifies } \mathbf{x} \text{ as True}.
\]

\end{theorem}

\emph{Proof idea.}
In a compatible fibred model, the accessibility relations between worlds
mirror the behavior of network instances on their inputs (condition C1),
and the fibring functions on worlds mirror the way
a parent network delegates part of its computation to children (condition C2).
Thus evaluating the fibred network on $\mathbf{x}$
corresponds exactly to checking the truth value of the formula
$\psi(\varphi,\mathcal{F})$ at the matching world,
since both follow the same recursive structure
given in Definition~\ref{def:syntactic-fragment}.

\emph{Proof.}
Fix a network instance $\mathcal{N}$ of the root architecture. Define the \emph{characteristic formula}
of $\mathcal{N}$ by
\[
\varphi \;:=\; \bigvee_{\mathbf{h}\in\{0,1\}^n:\ \mathcal{N}(\mathbf{h})>0}
\Big(\ \bigwedge_{h_k=1} p_k \ \wedge\ \bigwedge_{h_k=0} \neg p_k\ \Big).
\]

Fix $\mathcal{F} \in \mathscr{F}$, $\mathbf{x} \in\{0,1\}^n$ and $\mathcal{M}$ compatible with $\langle \mathcal{N},\mathcal{F}\rangle$ and $\mathbf{x}$. We prove by induction on the depth of $\mathcal{F}$ that, for the world
$w = (\pi_{u,1})^{-1}(\mathbf{x})$ at the root:
$
\mathcal{M},w \models \psi(\varphi,\mathcal{F})
\text{ \textit{iff} } \langle \mathcal{N},\mathcal{F}, \Tilde{f}\rangle \text{ classifies }\mathbf{x}\text{ as true}.
$

\emph{Base case (leaf).} If $\mathcal{F}$ has only the root then
$\psi(\varphi,\mathcal{F})=\varphi$ and by (C0), at the root model $\mathbf{m}_{u,1}$, $\varphi$ satisfies:
$\mathbf{m}_{u,1}, {\pi_{u,1}}^{-1}(\mathbf{h}) \models \varphi
\text{ \textit{iff} } \mathcal{N}(\mathbf{h})>0, \mathbf{h}\in\{0,1\}^n.$

\emph{Induction step.}
Suppose the root has children $u_1,\ldots,u_k$ with labels
$(\ell_i,S_i)$, subtrees $\mathcal{F}_i$ and $\Tilde{f}_i$ restrictions of $f$ to $\mathcal{F}_i$.
Run the root until layer $\ell_1$, obtain $\mathbf{x}_1$; fibre to $u_1$ by
$(\mathcal{N}_1,\mathbf{y}_1)=\Tilde{f}_{(u,u_1)} (\mathbf{x}_1)$; run the child fibred network
$\langle \mathcal{N}_1,\mathcal{F}_1, \Tilde{f}_1 \rangle$ on $\mathbf{y}_1$; splice its
output back to get $\mathbf{h}_1$; and continue similarly for $i=2,\dots,k$.
By (C1)–(C2) and the semantics of $\Box_{v,\ell}$, evaluating
$\psi(\varphi,\mathcal{F})=
\bigwedge_i \Box_{{v_i},\ell_i}\psi(\varphi,\mathcal{F}_i)$ at $w$
amounts to evaluating each $\psi(\varphi,\mathcal{F}_i)$ at the world reached
by the corresponding fibring jump. By the induction hypothesis, each of those
subformulas is true exactly when the corresponding child computation returns
the vector that is spliced into the parent. Hence
$\mathcal{M},w \models \psi(\varphi,\mathcal{F})$ iff the overall fibred
computation yields a final root output $>0$, i.e., iff
$\langle \mathcal{N},\mathcal{F}, \Tilde{f}\rangle$ classifies $\mathbf{x}$ as true. \qedsymbol



\section{Application to non-uniform expressiveness of GNNs, GATs and Transformer encoders}\label{sec:expressivity}

\paragraph{Graph Neural Networks.}
A Graph Neural Network (GNN) for  $\mathbb{A}$ is a tuple 
$\mathcal{G} = \langle \{\mathbf{A}^{\ell}\}_{\ell=1}^L, \{\mathbf{B}^{\ell}\}_{\ell=1}^L, \{\mathbf{b}^{\ell}\}_{\ell=1}^L \rangle$. 
For each layer $\ell$,  matrices  $\mathbf{A}^{\ell} \in \mathbb{Q}^{d_{\ell} \times d_{\ell-1}}$ and $\mathbf{B}^{\ell} \in \mathbb{Q}^{d_{\ell} \times d_{\ell-1}}$ are weight matrices and vector $\mathbf{b}^{\ell} \in \mathbb{Q}^{d_{\ell}}$ is a bias vector.
A Graph Attention Network (GAT) $\mathcal{G}'$ for $\mathbb{A}$ contains all the elements of a GNN plus an additional collection  $\{\mathbf{a}^{\ell}\}_{\ell=1}^L$ of attention vectors, where $\mathbf{a}^{\ell}\in \mathbb{Q}^{2 \cdot d_{\ell}}$. 
Both GNNs and GATs apply to undirected graphs $G = (V, E)$ where
a node vector $\mathbf{x}_u \in \{0,1\}^{d_0}$ is associated to each node $u \in V$ and where  the neighborhood of node $u$ is defined as 
$\mathrm{N} (u) = \{ v \in V : \{u,v\} \in E \}$. Their application to  graph $G$ with node vectors $\bm{x} := \left(\mathbf{x}_u \right)_{u \in V}$ yields a sequence  
$\left(\mathbf{x}^1_u \right)_{u \in V}, \ldots, \left(\mathbf{x}^L_u \right)_{u \in V}$ of node vectors and the final result is a graph with the same nodes and edges as $G$ but with updated node vectors $\{\mathbf{x}^L_u\}_{u \in V}$. When there is no ambiguity, we will refer to $\mathbf{x}^L_u$ as $\mathcal{G} (G, \bm{x}, u)$, the result at node $u$ of applying GNN (or GAT) $\mathcal{G}$ to graph $G$ with node features $\bm{x}$. 

In the application of GNN $\mathcal{G}$ 
to $G, \bm{x}$ is defined, for each $u \in V$, as 
$\mathbf{x}^{\ell}_u = \sigma^{\ell} (\mathbf{h}^{\ell}_u)$, where $\mathbf{x}^0_u = \mathbf{x}_u$ and
$
\label{eq:update}
\mathbf{h}^{\ell}_u    =     \mathbf{B}^{\ell} \cdot \mathbf{x}^{\ell-1}_u + \sum_{v \in \mathrm{N} (u)} \mathbf{A}^{\ell} \cdot \mathbf{x}^{\ell-1}_v + \mathbf{b}^{\ell}.
$
The application of GAT $\mathcal{G}'$ to $G$
is defined analogously by replacing this expression with
$\label{eq:update-GAT}
\mathbf{h}^{\ell}_u   =    \alpha_{u u}^\ell \cdot \mathbf{B}^{\ell} \cdot \mathbf{x}^{\ell-1}_u + \sum_{v \in \mathrm{N} (u)} \alpha_{u v}^\ell \cdot \mathbf{A}^{\ell} \cdot \mathbf{x}^{\ell-1}_v + \mathbf{b}^{\ell}
$,
where the \emph{attention coefficient} $\alpha_{u v}^\ell$ is the component associated to node $v$ in the following vector:
\begin{align}\label{eq:hardmax}
\mathrm{hardmax} \{  {\mathbf{a}^\ell}^T \cdot (\mathbf{A}^\ell \cdot \mathbf{x}_w^{\ell -1} || \mathbf{B}^\ell \cdot \mathbf{x}_u^{\ell -1}) \}_{w \in \mathrm{N} (u) \cup \{u\}}.&&
\end{align}
The hardmax function applied to a vector sets all occurrences of its largest value to the inverse of the number of its occurrences in the vector and all remaining components to $0$.
The expression within the hardmax function is a vector with a component for each node $w \in \mathrm{N} (u) \cup \{u\}$ and where each component is computed as the dot product of two vectors. 



\paragraph{Transformer Encoders with Hard Attention.} 
A Transformer encoder $\mathcal{T}$ for architecture $\mathbb{A}$ is 
defined similarly to a GAT as  
$\mathcal{T} = \langle \{\mathbf{A}^{\ell}\}_{\ell=1}^L, \{\mathbf{B}^{\ell}\}_{\ell=1}^L, \{\mathbf{b}^{\ell}\}_{\ell=1}^L, \{\mathbf{a}^{\ell}\}_{\ell=1}^L\rangle$ but is applicable only to specific types of input. While GATs are applied to arbitrary graphs, Transformer encoders are restricted to complete graphs, where each node is connected to every other node \citep{DBLP:journals/corr/abs-2104-13478}. Additionally, the input graph for $\mathcal{T}$ is constructed from an ordered sequence of tokens, with each token assigned a vector representation, which is then combined with a positional encoding vector to construct a token feature.
More precisely, for a complete graph with $s$ nodes associated to a token sequence $S$ of length $s$, each node is uniquely associated to a token $\xi$ at position $t \in \{0, ..., s-1\}$ in $S$ and each token is associated to a token feature 
$\mathbf{x}_{\xi}$ which is obtained as the sum
$\mathbf{x}_{\xi} = \mathit{vec}(\xi) + \mathit{pos}(t,s)$, where $\mathit{vec}$ is a vector representation function mapping tokens to vectors in $\{0,1\}^{d_0}$ and $\mathit{pos}$ is a positional encoding function mapping pairs of positions and sequence length to vectors in $\mathbb{Q}^{d_0}$.
The application of $\mathcal{T}$ to $S$ and token features $\bm{x} := \left( \mathbf{x}_\xi\right)_{\xi \in S}$ is then
obtained by applying expressions  \eqref{eq:update-GAT} and \eqref{eq:hardmax} to the complete graph with node features $\{\mathbf{x}_{\xi}\}_{\xi \in S}$. We will also note $\mathcal{T} (S, \bm{x}, \xi)$, the result at token $\xi$ of applying Transformer encoder $\mathcal{T}$ to sequence $S$ with token features $(\mathbf{x}_\xi)_{\xi \in S}$, i.e.\ $\mathcal{T} (S, \bm{x}, \xi)$ is the application of Transformer encoder $\mathcal{T}$ to $(S, \bm{x}, \xi)$.

\emph{Remark:} In this paper, we restrict ourselves to GNNs and GATs with local sum aggregation, rational coefficients, and applied to Boolean vectors. Our technical results also assume that each $\sigma^{\ell}$ is the truncated ReLU function mapping each component $x_i$ of $\mathbf{x}$ to $\mathrm{min} (1, \mathrm{max} (0,x_i))$. We also consider GATs and Transformers with hard attention only. To avoid repetitions, the technical results in this section will be stated for GATs, but they easily extend to GNNs and Transformer encoders, by considering that a GNN is a GAT without attention, and a Transformer encoder is a GAT on complete graphs with positional encodings.

If the output dimension of the GNN or GAT (respectively, Transformer encoder) is $1$, we can interpret them as \emph{node} (respectively, \emph{token}) \emph{classifiers}:
on input $(G, \bm{x})$ (respectively, $(S, \bm{x})$), for each node $u$ (respectively, for each token $\xi$), it outputs \emph{true} if $\mathbf{x}_u^L > 0$ (respectively, $\mathbf{x}_\xi^L > 0$) and \emph{false} otherwise. By abuse of notation, we sometimes write $\mathcal{G} (G, \bm{x}, u) = \mathsf{true}$ (or $\mathsf{false}$) and the same for $\mathcal{T} (S, \bm{x}, \xi)$.

\subsection{Fibred neural networks can non-uniformly describe GNNs, GATs and Transformer encoders}

The following result establishes that fibred neural networks provide a non-uniform description of GATs, i.e.\ given a GAT $\mathcal{G}$, for each input $G, u, \bm{x}$ there is a different fibred neural network whose computation coincides with that of $\mathcal{G}$. Interestingly, for a given input $(G, u)$, the fibring architecture is the same, and only the fibring functions vary for different node features $\bm{x}$.  This property is key to derive the (non-uniform) logical expressiveness result.

\begin{theorem}\label{gatasfnn}
For every tuple $\tau = \langle \mathcal{G}, G, u \rangle$, where $\mathcal{G}$ is a GAT with input/output dimensions $n/1$, $G = (V, E)$ a graph,  and $u \in V$, there exist a fibring architecture $\mathcal{F}^\tau$, a network instance $\mathcal{N}^{\tau}$ of the root architecture and a family $ \mathscr{G}_{\tau} = \{\Tilde{f}^\tau_{\bm{x}}\}_{\bm{x}}$ of collections of fibring functions matching $\mathcal{F}^\tau$, indexed by possible node features $\bm{x} = (\mathbf{x}_v)_{v \in V}$, 
 such that, for all $\bm{x}$, the fibred neural network $\langle \mathcal{N}^\tau, \mathcal{F}^\tau, \Tilde{f}^\tau_{\bm{x}}\rangle$ applied to $\mathbf{x}_u$  computes $\mathcal{G} (G, \bm{x}, u)$.
 \end{theorem}

\emph{Proof idea.} For each tuple $\tau = \langle \mathcal{G}, G, u \rangle$, the tree structure of the corresponding fibring architecture is given by the unraveling tree of node $u$ in $G$ at the depth of $\mathcal{G}$. Indeed, a GAT computes recursively w.r.t.\ the depth of its unraveling tree, as do fibred neural networks w.r.t.\ the depth of their fibring architecture. At each node, the fibring function returns the node features of the neighboring nodes and assigns the weights of the relevant layer of $\mathcal{G}$. In the computation of the fibred neural network, the step of "replacing the entries in $S_i$ of the vector $\mathbf{x}_i$ 
by $\langle \mathcal{N}_i,\mathcal{F}_i, \Tilde{f}_i \rangle(\mathbf{y}_i)$" is used to concatenate the outputs of the computations of the previous layer. The concatenation of outputs at the previous layer is then aggregated (or the attention layer is computed), using the relevant weights (or attention vectors) assigned by the fibring function upstream of the concatenation. The detailed proof is provided in the \Cref{appendix:proof}.

The same result holds for GNNs and Transformer encoders simply by removing the nodes and edges in the fibring architecture $\mathcal{F}^\tau$ that encode the GAT attention mechanism and from the fact that a Transformer encoder is a GAT applied to a complete graph with positional encoding incorporated into token features $(\mathbf{x}^0_\xi)_{\xi \in S}$.

\subsection{GNNs, GATs and Transformer encoders as fragments of fibred logics}



The following result establishes a non-uniform expressiveness result for GATs, characterized by a countably infinite family of formulas in the fragment of the fibred logic of interest.


\begin{theorem}\label{gatasfragment}

    Let $\tau = \langle \mathcal{G}, G, u \rangle$, $\mathcal{N}^{\tau}$, $\mathcal{F}^{\tau}$ and  $\mathscr{G}_{\tau} = \{\Tilde{f}^\tau_{\bm{x}}\}_{\bm{x}}$ be as in Theorem  \ref{gatasfnn}. Denote $u^\tau$ the root of $\mathcal{F}^{\tau}$. There exists a formula $\tilde{\varphi}^\tau$ in $\mathcal{L}_{\mathcal{F}^{\tau}}$ such that for each $\bm{x} = (\mathbf{x}_u)_{u \in V}$, for each fibred model $\mathcal{M}$ compatible with $\langle \mathcal{N}^\tau, \mathcal{F}^\tau, \Tilde{f}^\tau_{\bm{x}} \rangle$ and $\mathbf{x}_u$, the following holds:

    \[
    \mathcal{M}, (\pi_{u^\tau,1})^{-1} (\mathbf{x}_u) \models \Tilde{\varphi}^\tau \quad \text{iff}  \quad \mathcal{G} ( G, \bm{x}, u) = \mathsf{true} 
    \]
    
   
\end{theorem}

\emph{Proof idea.} It suffices to consider the fibred neural networks which reproduce the computations of $\mathcal{G}$ and to invoke the correspondence between fibred neural networks and our fibred logic. Since the formulas in our fibred logic only depend on the network at the root, and the fibring architecture, it follows that the formula does not depend on the node features.

\emph{Proof.} Let $\tau = \langle \mathcal{G}, G, u \rangle$ be a tuple with $\mathcal{G}$ a GAT with input/output dimensions $n/1$, $G$ a graph and $u$ a node in $G$. For each $\bm{x}$ node features for $G$, consider a fibred neural network $\Tilde{\mathcal{N}}^\tau_{\bm{x}} = \langle \mathcal{N}^\tau, \mathcal{F}^\tau, \Tilde{f}^\tau_{\bm{x}}\rangle$ obtained from Theorem \ref{gatasfnn}.

    By Theorem \ref{fnneqlogic} applied to $\mathcal{N}^\tau$ and $\mathcal{F}^\tau$, there exists a propositional formula $\varphi^\tau$ such that for each fibred model $\mathcal{M}$ compatible with $\Tilde{\mathcal{N}}^\tau_{\bm{x}}$ and $\mathbf{x}_u$, $\mathcal{M}, (\pi_{u^\tau,1})^{-1} (\mathbf{x}_u) \models \psi (\varphi^{\tau}, \mathcal{F}^\tau)$ iff $\Tilde{\mathcal{N}}^\tau_{\bm{x}} (\mathbf{x}_u) > 0$. Furthermore by Theorem \ref{gatasfnn} $\Tilde{\mathcal{N}}^\tau_{\bm{x}} (\mathbf{x}_u) > 0$ iff $\mathcal{G} (G, \bm{x}, u) = \mathsf{true}$, which gives us the result. \qedsymbol

The same result also holds for GNNs and Transformer encoders by replacing the fibring architecture $\mathcal{F}^\tau$ obtained from Theorem \ref{gatasfnn}. In particular, to incorporate positional encodings for Transformers, it suffices to replace $\pi_{u,1}$ by the bijection $\pi_{\xi, 1}$ from worlds in $\mathbf{m}_{\xi, 1}$ to $\{pos(t, s), 1 + pos(t,s)\}^n$ where $t$ is the position of token $\xi$ and $s$ is the length of the sequence.

\section{Discussion, Conclusion and Future Work}\label{sec:discussion}

For a given instance $\mathcal{G}$ of a neural architecture (a GNN, a GAT or a Transformer encoder), our non-uniform expressiveness results provide a countable family of formulas that characterize the network instance. 
We note that the result of \citet{grau2025correspondenceboundedgraphneural} on bounded GNNs also implies a non-uniform expressiveness result in terms of first-order logic formulas: to construct a countable family of formulas that characterize a GNN $\mathcal{G}$, we can for example enumerate for $k \in \mathbb{N}$ the first-order logic formula of the bounded GNN obtained by applying $\mathcal{G}$ to all graphs with degree $k$. 

Closing the gap to uniform expressiveness requires collapsing, for each instance $\mathcal{G}$, the corresponding family into a single formula in another logic. With the uniform expressiveness result of \citet{benedikt_et_al:LIPIcs.ICALP.2024.127}, we already know that, for GNNs, the family must collapse to a Presburger formula. The problem remains open for GATs and Transformer encoders. In particular, for Transformer encoders only a lower bound of the resulting logic is known \citep{barcelo2024logical}. 

We now present ideas towards the unification of uniform expresiveness results using fibring. Consider the countable family of fibred neural networks 
characterizing a GNN, given in the proof of \Cref{gatasfnn}. These fibred neural networks follow a common recursive structure, and by applying them to all possible input vectors $\mathbf{x} \in \{0,1\}^n$, we can collect at each component (Kripke models of compatible fibred models) the vectors attained by the computations of the fibred neural networks, following the procedure in the proof of \Cref{prop:class-compatible}. Given the correspondence between the structure of the fibring architectures and the layers of the GNN (as seen in the proof of \Cref{gatasfnn}), the set of vectors collected by this procedure is closely related to the \emph{$\ell$-spectrum} of the GNN, defined in \citet{benedikt_et_al:LIPIcs.ICALP.2024.127} as the set of all vectors attainable by the GNN at the $\ell$-th layer when applied to all possible inputs. In particular, the finiteness of the $\ell$-spectrum is a key argument used in the derivation of their uniform expressiveness results. An idea would thus be to identify commonalities between the fibred formulas for different inputs associated to the same GNN, in order to map them to the single Presburger formula. Many technical details must be made precise to close this gap, and we leave this research as future work. We believe, however, that the definition of fibring provided here will help systematize this investigation. Repeating the exercise to find commonalities in the fibred formulas for GATs and Transformer encoders constitutes, in this way, a new approach towards deriving uniform expressiveness results. The fibring formalism is expected to provide a unified lens, although non-uniform at this point, on the logical expressiveness of different network architectures, and might even hold the promise of offering a common methodology to uncover expressiveness results in the future.

The intuition of fibred logics applied to neural architectures, viewing neural networks as combining underlying logics, is appealing and similar in spirit to the current trend in interpretability research which tries to break down network computations into several modular reasoning steps, e.g.\citep{AmeisenEtAl2025CircuitTracing, he2025towards}. We believe that scrutinizing the commonalities between fibred formulas for typical inputs, thereby reverse-engineering the fibred logical theories learned by neural networks, constitutes an exciting new take on the future possible extraction of interpretable logical rules capturing neural network's most relevant computations. Therefore, we posit that our approach to logical expressiveness via fibring may share common ground with interpretability research and hope that this paper will serve as a foundation in this direction. Furthermore, achieving uniform expressiveness results using the fibring formalism could enable new results on formal verification by leveraging existing results on the complexity of modal and fibred logics, e.g.\ \citep{WuEtAl2015NPcompleteFibringLogic}. 

This paper is the first to establish the exact correspondence between fibring neural networks and fibring modal logics. The formalism has allowed us to derive non-uniform logical expressiveness results for several modern neural architectures. A gap remains unresolved to bridge non-uniform and uniform expressiveness results, and we hope our work will inspire further research in this direction. We think that fibring as a formalism has the potential to enable the unification of expressiveness results for various network architectures, which were initially derived using different methodologies, with future applications in interpretability and verification.

\bibliography{example_paper}

\begin{thebibliography}{24}
\providecommand{\natexlab}[1]{#1}
\providecommand{\url}[1]{\texttt{#1}}
\expandafter\ifx\csname urlstyle\endcsname\relax
  \providecommand{\doi}[1]{doi: #1}\else
  \providecommand{\doi}{doi: \begingroup \urlstyle{rm}\Url}\fi

\bibitem[Ameisen et~al.(2025)Ameisen, Lindsey, Pearce, Gurnee, Turner, Chen, Citro, Abrahams, Carter, Hosmer, Marcus, Sklar, Templeton, Bricken, McDougall, Cunningham, Henighan, Jermyn, Jones, Persic, Qi, Thompson, Zimmerman, Rivoire, Conerly, Olah, and Batson]{AmeisenEtAl2025CircuitTracing}
Emmanuel Ameisen, Jack Lindsey, Adam Pearce, Wes Gurnee, Nicholas~L. Turner, Brian Chen, Craig Citro, David Abrahams, Shan Carter, Basil Hosmer, Jonathan Marcus, Michael Sklar, Adly Templeton, Trenton Bricken, Callum McDougall, Hoagy Cunningham, Thomas Henighan, Adam Jermyn, Andy Jones, Andrew Persic, Zhenyi Qi, T.~Ben Thompson, Sam Zimmerman, Kelley Rivoire, Thomas Conerly, Chris Olah, and Joshua Batson.
\newblock Circuit tracing: Revealing computational graphs in language models.
\newblock Anthropic / Transformer-Circuits Blog preprint, March 27 2025.
\newblock URL: \url{https://transformer-circuits.pub/2025/attribution-graphs/methods.html}.

\bibitem[Barcel{\'o} et~al.(2020)Barcel{\'o}, Kostylev, Monet, P{\'e}rez, Reutter, and Silva]{barcelo2020logical}
Pablo Barcel{\'o}, Egor~V. Kostylev, Mika{\"e}l Monet, Jorge P{\'e}rez, Juan~L. Reutter, and Juan~Pablo Silva.
\newblock The logical expressiveness of graph neural networks.
\newblock In \emph{8th International Conference on Learning Representations (ICLR)}, 2020.

\bibitem[Barcelo et~al.(2024)Barcelo, Kozachinskiy, Lin, and Podolskii]{barcelo2024logical}
Pablo Barcelo, Alexander Kozachinskiy, Anthony~Widjaja Lin, and Vladimir Podolskii.
\newblock Logical languages accepted by transformer encoders with hard attention.
\newblock In \emph{The Twelfth International Conference on Learning Representations}, 2024.
\newblock URL \url{https://openreview.net/forum?id=gbrHZq07mq}.

\bibitem[Benedikt et~al.(2024)Benedikt, Lu, Motik, and Tan]{benedikt_et_al:LIPIcs.ICALP.2024.127}
Michael Benedikt, Chia-Hsuan Lu, Boris Motik, and Tony Tan.
\newblock {Decidability of Graph Neural Networks via Logical Characterizations}.
\newblock In Karl Bringmann, Martin Grohe, Gabriele Puppis, and Ola Svensson (eds.), \emph{51st International Colloquium on Automata, Languages, and Programming (ICALP 2024)}, volume 297 of \emph{Leibniz International Proceedings in Informatics (LIPIcs)}, pp.\  127:1--127:20, Dagstuhl, Germany, 2024. Schloss Dagstuhl -- Leibniz-Zentrum f{\"u}r Informatik.
\newblock ISBN 978-3-95977-322-5.
\newblock \doi{10.4230/LIPIcs.ICALP.2024.127}.
\newblock URL \url{https://drops.dagstuhl.de/entities/document/10.4230/LIPIcs.ICALP.2024.127}.

\bibitem[Besold et~al.(2022)Besold, d'Avila Garcez, Bader, Bowman, Domingos, Hitzler, K\"{u}hnberger, Lamb, Machado Vieira~Lima, de~Penning, Pinkas, Poon, and Zaverucha]{Besold2022}
Tarek~R. Besold, Artur d'Avila Garcez, Sebastian Bader, Howard Bowman, Pedro Domingos, Pascal Hitzler, Kai-Uwe K\"{u}hnberger, Lu\'{\i}s~C. Lamb, Priscila Machado Vieira~Lima, Leo de~Penning, Gadi Pinkas, Hoifung Poon, and Gerson Zaverucha.
\newblock \emph{Neural-Symbolic Learning and Reasoning: A Survey and Interpretation}, volume 342 of \emph{Frontiers in Artificial Intelligence and Applications}, chapter~1, pp.\  1--51.
\newblock IOS Press, 2022.

\bibitem[Bronstein et~al.(2021)Bronstein, Bruna, Cohen, and Velickovic]{DBLP:journals/corr/abs-2104-13478}
Michael~M. Bronstein, Joan Bruna, Taco Cohen, and Petar Velickovic.
\newblock Geometric deep learning: Grids, groups, graphs, geodesics, and gauges.
\newblock \emph{CoRR}, abs/2104.13478, 2021.
\newblock URL \url{https://arxiv.org/abs/2104.13478}.

\bibitem[Brown et~al.(2020)Brown, Mann, Ryder, Subbiah, Kaplan, Dhariwal, Neelakantan, Shyam, Sastry, Askell, et~al.]{brown2020language}
Tom~B. Brown, Benjamin Mann, Nick Ryder, Melanie Subbiah, Jared~D. Kaplan, Prafulla Dhariwal, Arvind Neelakantan, Pranav Shyam, Girish Sastry, Amanda Askell, et~al.
\newblock Language models are few-shot learners.
\newblock In \emph{Advances in Neural Information Processing Systems}, volume~33, pp.\  1877--1901, 2020.

\bibitem[Chellas(1980)]{chellas1980modal}
Brian~F. Chellas.
\newblock \emph{Modal Logic: An Introduction}.
\newblock Cambridge University Press, Cambridge, 1980.
\newblock ISBN 9780521295154.

\bibitem[Cuenca~Grau et~al.(2025)Cuenca~Grau, Feng, and Walega]{grau2025correspondenceboundedgraphneural}
Bernardo Cuenca~Grau, Eva Feng, and Przemyslaw~A. Walega.
\newblock The correspondence between bounded graph neural networks and fragments of first-order logic, 2025.
\newblock URL \url{https://arxiv.org/abs/2505.08021}.

\bibitem[Devlin et~al.(2019)Devlin, Chang, Lee, and Toutanova]{devlin2019bert}
Jacob Devlin, Ming-Wei Chang, Kenton Lee, and Kristina Toutanova.
\newblock {BERT}: Pre-training of deep bidirectional transformers for language understanding.
\newblock In \emph{Proceedings of the 2019 Conference of the North American Chapter of the Association for Computational Linguistics: Human Language Technologies, Volume 1 (Long and Short Papers)}, pp.\  4171--4186. Association for Computational Linguistics, 2019.

\bibitem[d’Avila Garcez \& Lamb(2023)d’Avila Garcez and Lamb]{GarcezLamb2023}
Artur d’Avila Garcez and Luís~C. Lamb.
\newblock Neurosymbolic {AI}: the 3rd wave.
\newblock \emph{Artificial Intelligence Review}, 56\penalty0 (11):\penalty0 12387--12406, November 2023.
\newblock \doi{10.1007/s10462-023-10448-w}.
\newblock URL \url{https://doi.org/10.1007/s10462-023-10448-w}.

\bibitem[Gabbay(1999)]{Gabbay1999-GABFL}
Dov~M. Gabbay.
\newblock \emph{Fibring Logics}.
\newblock Clarendon Press, New York, 1999.

\bibitem[Garcez \& Gabbay(2004)Garcez and Gabbay]{10.5555/1597148.1597205}
Artur S.~d'Avila Garcez and Dov~M. Gabbay.
\newblock Fibring neural networks.
\newblock In \emph{Proceedings of the 19th National Conference on Artifical Intelligence}, AAAI'04, pp.\  342–347. AAAI Press, 2004.
\newblock ISBN 0262511835.

\bibitem[Grohe(2023)]{Grohe2023}
Martin Grohe.
\newblock The descriptive complexity of graph neural networks.
\newblock In \emph{38th ACM/IEEE Symposium on Logic in Computer Science (LICS)}, pp.\  1--14, 2023.
\newblock \doi{10.1109/LICS56636.2023.10175735}.

\bibitem[Hao et~al.(2022)Hao, Angluin, and Frank]{hao2022formal}
Yiding Hao, Dana Angluin, and Robert Frank.
\newblock Formal language recognition by hard attention transformers: Perspectives from circuit complexity.
\newblock \emph{Transactions of the Association for Computational Linguistics}, 10:\penalty0 800--810, 2022.

\bibitem[He et~al.(2025)He, Zheng, Dong, Zhu, Chen, and Li]{he2025towards}
Yinhan He, Wendy Zheng, Yushun Dong, Yaochen Zhu, Chen Chen, and Jundong Li.
\newblock Towards global-level mechanistic interpretability: A perspective of modular circuits of large language models.
\newblock In \emph{Forty-second International Conference on Machine Learning}, 2025.
\newblock URL \url{https://openreview.net/forum?id=do5vVfKEXZ}.

\bibitem[Lamb et~al.(2020)Lamb, d'Avila Garcez, Gori, Prates, Avelar, and Vardi]{Lamb2020}
Lu\'{\i}s~C. Lamb, Artur d'Avila Garcez, Marco Gori, Marcelo O.~R. Prates, Pedro H.~C. Avelar, and Moshe~Y. Vardi.
\newblock Graph neural networks meet neural-symbolic computing: A survey and perspective.
\newblock In \emph{Proceedings of the 29th International Joint Conference on Artificial Intelligence (IJCAI)}, pp.\  4877--4884, 7 2020.
\newblock \doi{10.24963/ijcai.2020/679}.
\newblock Survey track.

\bibitem[Nunn et~al.(2024)Nunn, Sälzer, Schwarzentruber, and Troquard]{ijcai2024p391}
Pierre Nunn, Marco Sälzer, François Schwarzentruber, and Nicolas Troquard.
\newblock A logic for reasoning about aggregate-combine graph neural networks.
\newblock In Kate Larson (ed.), \emph{Proceedings of the Thirty-Third International Joint Conference on Artificial Intelligence, {IJCAI-24}}, pp.\  3532--3540. International Joint Conferences on Artificial Intelligence Organization, 8 2024.
\newblock \doi{10.24963/ijcai.2024/391}.
\newblock URL \url{https://doi.org/10.24963/ijcai.2024/391}.
\newblock Main Track.

\bibitem[OpenAI(2023)]{openai2023gpt4}
OpenAI.
\newblock {GPT-4} technical report.
\newblock \emph{arXiv preprint arXiv:2303.08774}, 2023.
\newblock URL \url{https://arxiv.org/abs/2303.08774}.

\bibitem[Salamat et~al.(2021)Salamat, Luo, and Jafari]{SALAMAT2021106817}
Amirreza Salamat, Xiao Luo, and Ali Jafari.
\newblock Heterographrec: A heterogeneous graph-based neural networks for social recommendations.
\newblock \emph{Knowledge-Based Systems}, 217:\penalty0 106817, 2021.
\newblock ISSN 0950-7051.
\newblock \doi{https://doi.org/10.1016/j.knosys.2021.106817}.
\newblock URL \url{https://www.sciencedirect.com/science/article/pii/S0950705121000800}.

\bibitem[Vaswani et~al.(2017)Vaswani, Shazeer, Parmar, Uszkoreit, Jones, Gomez, Kaiser, and Polosukhin]{NIPS2017_3f5ee243}
Ashish Vaswani, Noam Shazeer, Niki Parmar, Jakob Uszkoreit, Llion Jones, Aidan~N Gomez, \L~ukasz Kaiser, and Illia Polosukhin.
\newblock Attention is all you need.
\newblock In I.~Guyon, U.~Von Luxburg, S.~Bengio, H.~Wallach, R.~Fergus, S.~Vishwanathan, and R.~Garnett (eds.), \emph{Advances in Neural Information Processing Systems}, volume~30. Curran Associates, Inc., 2017.
\newblock URL \url{https://proceedings.neurips.cc/paper_files/paper/2017/file/3f5ee243547dee91fbd053c1c4a845aa-Paper.pdf}.

\bibitem[Wu et~al.(2015)Wu, Jiang, Huang, Chao, and Zhou]{WuEtAl2015NPcompleteFibringLogic}
Yin Wu, Min Jiang, Zhongqiang Huang, Fei Chao, and Changle Zhou.
\newblock An np‐complete fragment of fibring logic.
\newblock \emph{Annals of Mathematics and Artificial Intelligence}, 75\penalty0 (3-4):\penalty0 391--417, 2015.
\newblock \doi{10.1007/s10472-015-9468-4}.

\bibitem[Xiong et~al.(2021)Xiong, Xiong, Chen, Jiang, and Zheng]{xiong2021graph}
Jiacheng Xiong, Zhaoping Xiong, Kaixian Chen, Hualiang Jiang, and Mingyue Zheng.
\newblock Graph neural networks for automated de novo drug design.
\newblock \emph{Drug Discovery Today}, 26\penalty0 (6):\penalty0 1382--1393, 2021.

\bibitem[Ye et~al.(2022)Ye, Kumar, Ong~Sing, Song, and Wang]{ye2022comprehensive}
Zi~Ye, Yogan~Jaya Kumar, Goh Ong~Sing, Fengyan Song, and Junsong Wang.
\newblock A comprehensive survey of graph neural networks for knowledge graphs.
\newblock \emph{IEEE Access}, 10:\penalty0 75728--75747, 2022.

\end{thebibliography}
\bibliographystyle{iclr2026/iclr2026_conference}


\newpage
\appendix
\onecolumn

\section{Proof of \Cref{gatasfnn}}\label{appendix:proof}
    Let $\tau = \langle \mathcal{G}, G, u \rangle$ where $\mathcal{G} = \langle \{\mathbf{A}^{\ell}\}_{\ell=1}^L, \{\mathbf{B}^{\ell}\}_{\ell=1}^L, \{\mathbf{b}^{\ell}\}_{\ell=1}^L, \{\mathbf{a}^{\ell}\}_{\ell=1}^L\rangle$ is a GAT instance, $G = (V,E)$ is a graph with maximum degree $m$ and $u$ is a node in $G$.

    Define $\mathcal{N}^{\tau}$ as the two-layer linear network whose first layer is the concatenation of $m +1$ identity matrices (with no bias) and the second layer is the concatenation of the matrix $\mathbf{B}^L$, and $m$ times the matrix $\mathbf{A}^L$ (with bias $\mathbf{b}^L$).

    \begin{definition}
        Let \( G = (V, E)\) be a graph. A \emph{lazy walk} in $G$ is a finite sequence of nodes $(u_0,..., u_k)$ such that for each $i \in \{1,\ldots,k\}$ $u_{i-1} = u_i$ or $(u_{i-1}, u_i) \in E$. A lazy walk differs from a \emph{path} in that one can stay on the same node multiple times.
    \end{definition}
    
    \begin{definition} Let \( G = (V, E)\) be a graph, \( u \in V\), and \( L \in \mathbb{N} \). The (lazy) unravelling of node \( u \) in \( G \) at depth \( L \) is the graph that is the tree having:
    \begin{itemize}
        \item a root $u$
        \item a node \( (u, u_1, \dots, u_\ell) \) for each lazy walk \( (u, u_1, \dots, u_\ell) \) in \( G \) with \( 1 \leq \ell \leq L \), and
        \item an edge between \( (u, u_1, \dots, u_{\ell-1}) \) and \( (u, u_1, \dots, u_\ell) \) when $u_{\ell-1} = u_\ell$ or \( ( u_{\ell-1}, u_\ell) \in E \) (assuming that \( u_0 \) is \( u \)).
    \end{itemize}
    \end{definition}

     The fibring architecture $\mathcal{F}^{\tau}$ is constructed from the lazy unravelling of $u$ in $G$ at depth $L$ to which we add (in order to encode the attention mechanism) some nodes and edges in the following way:
    for each $\ell \in \{1 \ldots L \}$ (assuming that \( u_0 \) is \( u \)), each node $(u, u_1, ..., u_{\ell-1})$ has an extra child node $v_\ell$, and we require that each node $v_\ell$ has no child node. Note that by this construction, each node $(u, u_1, ..., u_\ell)$ has $| \mathrm{N} (u_{\ell}) | + 2$ children nodes (namely the nodes $(u, u_1, ..., u_\ell, w)$ for $w \in \{u_\ell\} \cup \mathrm{N} (u_{\ell})$, and the node $v_{\ell+1}$). 

    Let $\bm{x}$ be node features for $G$.
    
    The label of the edge between\( (u, u_1, \dots, u_{\ell-1}) \) and \( (u, u_1, \dots, u_\ell) \) is defined by layer number 1 and the set of positions designed so the final output vector is the concatenation of the children computations. Its fibring function is as follows: on input the node feature $\mathbf{x}^0_{u_\ell}$, it returns the concatenation of vectors $\mathbf{x}^{0}_w$ for $w \in \{u_\ell\} \cup \mathrm{N} (u_{\ell})$ and a network with the relevant weights of layer $L-\ell$ in $\mathcal{G}$ designed so that, in the computation of the fibred network, the network $\mathcal{N}_{(u, u_1, \dots, u_\ell)}$ applied to the concatenation of vectors $\mathbf{h}^{L-\ell+1}_w$ for $w \in \{u_\ell\} \cup \mathrm{N} (u_{\ell})$ returns the concatenation of $\mathbf{A}^{L-\ell} \cdot \sigma (\mathbf{h}_{w}^{L-\ell+1})$  (and $\mathbf{B}^{L-\ell} \cdot \sigma (\mathbf{h}_{w}^{L-\ell+1})$ when $w = u_\ell$).

    The label of the edge between \( (u, u_1, \dots, u_{\ell-1}) \) and $v_\ell$ is defined by layer number 2 and the set of positions is all positions of layer 2. Its fibring function takes as input a vector $\mathbf{y}$ and returns the same vector $\mathbf{y}$ (\emph{self-fibring}) and a network using the relevant attention vector $\mathbf{a}^{L-\ell}$ (and the bias) of layer $L-\ell$ designed so that when $\mathbf{y}$ is the concatenation of vectors $\mathbf{y}_{w}^{L-\ell}$ for $w \in \{u_\ell\} \cup \mathrm{N} (u_{\ell})$, then $\mathcal{N}_{v_\ell} (\mathbf{y})$ is the sum of vectors $\alpha^{L-\ell}_{u_\ell w} \cdot \mathbf{y}^{L-\ell}_w$ (plus the bias $\mathbf{b}^{L-\ell}$). 

    The neural architectures labelling nodes have the dimensions and nonlinearities required to realise the networks described above (in particular, the neural architectures labelling nodes $v_\ell$ uses the $\mathrm{hardmax}$ nonlinearity), and it is easy to see that the set of neural architectures labelling nodes, layer numbers, and sets of positions labelling edges are fully determined by $\tau$ (in particular, they do not depend on $\bm{x}$). 

    As a result, since $v_\ell$ is a leaf node, it follows that, in the computation of the fibred network, the output vector at node \( (u, u_1, \dots, u_{\ell-1}) \) is given by $\mathcal{N}_{v_\ell} (\mathbf{y})$ where $\mathbf{y}$ is the concatenation of the children computations, i.e.\ from nodes \( (u, u_1, \dots, u_{\ell-1}, u_\ell) \).

    We prove by (descending) induction on the depth that the fibred network $\Tilde{\mathcal{N}} = \langle \mathcal{N}^\tau, \mathcal{F}^\tau, \Tilde{f}^\tau_{\bm{x}} \rangle $ verifies $\Tilde{\mathcal{N}} (\mathbf{x}^0_u) = \mathbf{h}^L_u$.

    \emph{Base ($L-1$):} By construction, each child computation is given by the concatenation of $\mathbf{A}^{1} \cdot \mathbf{x}_{w}^{0}$  (and $\mathbf{B}^{1} \cdot \mathbf{x}_{w}^{0}$) for $w \in \{u_{L-1}\} \cup \mathrm{N} (u_{L-1})$, where $u_{L-1}$ is a neighbour reached at depth $L-1$ starting from the root $u$. Applying $\mathcal{N}_{v_L}$ to the concatenation of these children computations yields the sum of vectors $\alpha^{1}_{u_{L-1} w} \cdot \mathbf{A}^{1} \cdot \mathbf{x}_{w}^{0}$ (plus the bias $\mathbf{b}^1$), which is $\mathbf{h}^1_{u_{L-1}}$.

    \emph{Induction step ($\ell \to \ell -1$):} By inductive hypothesis, the output vector at each node \( (u, u_1, \dots, u_{\ell}) \) is $\mathbf{h}^{L-\ell}_{u_{\ell}}$ where $u_{\ell}$ is a neighbour reached at depth $\ell$ starting from the root $u$. Computing the output at node \( (u, u_1, \dots, u_{\ell-1}) \): by construction, each child computation is thus given by the concatenation of $\mathbf{A}^{L-\ell+1} \cdot \sigma (\mathbf{h}_{w}^{L-\ell})$  (and $\mathbf{B}^{L-\ell+1} \cdot \sigma (\mathbf{h}_{w}^{L-\ell})$) for $w \in \{u_{\ell-1}\} \cup \mathrm{N} (u_{\ell-1})$. Applying $\mathcal{N}_{v_\ell}$ to the concatenation of these children computations yields the sum of vectors $\alpha^{L-\ell+1}_{u_{\ell-1} \; w} \cdot \mathbf{A}^{L - \ell + 1} \cdot \mathbf{x}_{w}^{L-\ell}$ (plus the bias $\mathbf{b}^{L - \ell + 1}$), which is $\mathbf{h}^{L-\ell+1}_{u_{\ell-1}}$.

\end{document}